\documentclass{article}

\usepackage{microtype}
\usepackage{graphicx}
\usepackage{subfigure}
\usepackage{booktabs} % for professional tables
\usepackage{bm}      % for \bm macro

\usepackage{hyperref}

\usepackage[accepted]{icml2021}

\icmltitlerunning{Out of Distribution Detection and Adversarial Attacks on Deep Neural Networks for Robust Medical Image Analysis}

\begin{document}

\twocolumn[
\icmltitle{Out of Distribution Detection and Adversarial Attacks \\on Deep Neural Networks for Robust Medical Image Analysis}




\begin{icmlauthorlist}
\icmlauthor{Anisie Uwimana}{goo}
\icmlauthor{Ransalu Senanayake}{to}
\end{icmlauthorlist}

\icmlaffiliation{to}{Durand Building, 496 Lomita Mall, Stanford University, Stanford, CA 94305}
\icmlaffiliation{goo}{African Institute for Mathematical Sciences (AIMS), Rwanda, Kigali}

\icmlcorrespondingauthor{Anisie Uwimana}{auwimana@aimsammi.org}
% \icmlcorrespondingauthor{Ransalu Senanayake}{ransalu@stanford.edu}

\icmlkeywords{Machine Learning, ICML}

\vskip 0.3in
]



\printAffiliationsAndNotice{}  % leave blank if no need to mention equal contribution

\begin{abstract}
Deep learning models have become a popular choice for medical image analysis. However, the poor generalization performance of deep learning models limits them from being deployed in the real world as robustness is critical for medical applications. For instance, the state-of-the-art Convolutional Neural Networks (CNNs) fail to detect adversarial samples or samples drawn statistically far away from the training distribution. In this work, we experimentally evaluate the robustness of a Mahalanobis distance-based confidence score, a simple yet effective method for detecting abnormal input samples, in classifying malaria parasitized cells and uninfected cells. Results indicated that the Mahalanobis confidence score detector exhibits improved performance and robustness of deep learning models, and achieves state-of-the-art performance on both \emph{out-of-distribution (OOD)} and \emph{adversarial} samples.
\end{abstract}

\section{Introduction}

Deep learning is increasingly making its way into groundbreaking technologies that have high-value applications in the real-world clinical environment. Innovative medical imaging applications and diagnostics are among the most exciting use cases. One such application is developing microscopy-based malaria diagnosis procedures~\cite{ravendran2015moment,silva2013computer,yang2019deep}. Malaria is a deadly mosquito-borne disease infecting around $300$ million people annually \cite{WHO}. Since it is mostly prevalent in low-income countries, developing semi-automated microscopy techniques, as alternatives to polymerase chain reaction (PCR) tests and rapid diagnostic tests (RDT), is a low-cost and reliable solution \cite{wongsrichanalai2007review}.

\subsection{Applications of deep learning in medical diagnosis}

\citet{esteva2017dermatologist} developed a convolutional neural network (CNN) model that was trained on $130,000$ clinical images of skin pathologies to detect cancer. The proposed model achieves performance on par with all tested experts, demonstrating an artificial intelligence model capable of classifying skin cancer with a level of competence comparable to dermatologists. In $2018$, another research study showed that a convolutional neural network trained to analyze dermatology images identified melanoma with ten percent more specificity than human clinicians~\citep{haenssle2018man}. Another algorithm trained on $42,000$ chest CT scans outperformed expert radiologists in detecting lung cancers~\citep{ardila2019end}. It was able to find malignant lung modes $5$-$9.5\%$ more often than human specialists. Recently, a CNN model designed to predict malignancy and identify 134 skin disorders~\citep{cho2020dermatologist}. The proposed algorithm is capable of distinguishing, at the human expert level, melanoma from birthmarks. There have also been various studies on assessing the uncertainty and robustness in medical data~\cite{senanayake2016predicting,laves2020well,asgharnezhad2020objective}.

\subsection{Applications of deep learning for malaria diagnosis}

Various computer vision algorithms have been used for malaria diagnosis~\cite{ravendran2015moment}. Deep learning algorithms are recently being used increasingly by researchers especially for malaria detection because of its applicability in building automated diagnostic system. \citet{Liang2016CNNbasedIA} presented a 16-layer CNN towards classifying uninfected and parasitized cells. The study reported that the custom model was more accurate, sensitive, and specific than the pre-trained model. \citet{7897215} evaluated three well-known CNNs (LeNet, AlexNet and GoogLeNet) on  classifying parasite/not parasite slide images of thin blood stains. Simulation results showed that all three CNNs achieved classification accuracy scores of over $95\%$. \citet{rajaraman2018pre} introduced a pretrained CNN as a feature extractor towards improved malaria parasite detection in thin blood smear images, and the results present the use of pretrained CNNs as a promising tool in malaria detection.

\citet{Dipanjan} has demonstrated that deep neural networks can be used to detect malaria from microscopic images. However, in order to deploy such systems in medical facilities, it is vital to ensure that the automated detection systems are indeed robust. Nonetheless, deep learning algorithms work on the premise that both training and test data are drawn from the same application-specific distribution. However, in real-world applications, they need  to be able to robustly handle anomalous inputs including, 1) adversarial samples arising from image distortion and 2) samples drawn from a different distribution but belong to the same input space.

\begin{figure}[h]
\vskip 0.2in
\begin{center}
\centerline{\includegraphics[width=\columnwidth]{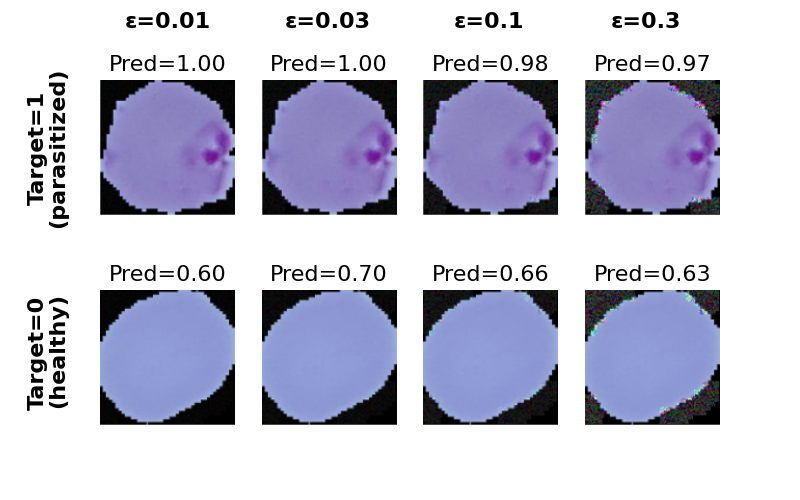}}
\caption{FGSM adversarial attack on parasitized and healthy cells for varying levels of noise. Noise can be visually inspected on the black areas and cell boundary, especially for $\epsilon=0.3$). On the top row, parasites can be seen in dark color inside the cell. The prediction probabilities are indicated above each image.}
\label{fig:1}
\end{center}
\vskip -0.2in
\end{figure}

In this work, we propose using a Mahalanobis distance-based confidence score method \cite{lee2018simple,kamoi2020mahalanobis,nitsch2020out} for detecting abnormal (both OOD and adversarial) samples to improve the performance and robustness of pre-trained convolutional neural network models to improve the robustness of malaria detection (Figure~\ref{fig:1}). The suggested method gives better results compared to the current state-of-the-art method ODIN \cite{liang2017enhancing} in detecting OOD malaria samples. We also demonstrate that Mahalanobis distance-based confidence score outperforms the state-of-the-art detection, LID, in all test cases, in detecting adversarial samples generated by four adversarial attacking methods: FGSM \cite{goodfellow2015explaining}, BIM \cite{kurakin2016adversarial}, DeepFool \cite{moosavi2016deepfool}, and CW \cite{carlini2017adversarial}.

\section{Robustness of deep learning models}

The robustness of deep learning algorithms needs to be evaluated before deploying them in real-wold. Therefore, it is crucial to ensure the neural networks can detect abnormal inputs in safety- and security-sensitive applications such as medical diagnosis, biometric authentication, intrusion detection, and autonomous driving~\citep{emmott2016metaanalysis,nitsch2020out}.

\subsection{Robust out-of-distribution detection for neural networks}

Out-of-Distribution (OOD) samples, the test samples that are not well covered by training data, is a major cause of poor performance in deep learning models. OOD samples are able to both evade the deep learning algorithms as well as achieve targeted misclassification with high confidence. There are currently many approaches that can detect OOD examples. They work well when tested on natural samples from a distribution that is sufficiently different from the distribution of the training data \cite{chen2020robust}.
 
\citet{hendrycks2016baseline} recently proposed a  baseline for detecting misclassified and OOD examples in deep neural networks (DNNs), and \citet{liang2017principled} improved  it by processing the input and output of the DNNs. The Softmax Baseline Mode computes softmax probabilities with the fast-growing exponential function. Thus minor changes to the softmax inputs, can lead to major changes in the output distribution. A softmax baseline method uses probabilities from softmax distributions to predict whether a test example is from a different distribution from the training data or from within the same distribution. \citet{liang2017enhancing} proposed ODIN (Out-of-DIstribution detector for Neural networks) which is a simple and effective method for detecting OOD images in neural networks. ODIN does not require re-training the neural network and is compatible with diverse network architectures and datasets.

\subsection{Robust adversarial detection for neural networks}
 
Recent studies have concentrated on identifying adversarial examples despite the inefficiency of adversarial defense \citep{NIPS2016_6331}. \citet{goodfellow2014generative} suggested a framework for estimating abnormal samples via adversarial networks. Local Intrinsic Dimensionality (LID) is one of the successful adversarial detection techniques proposed by \citet{ma2018characterizing}. With the assumption that adversarial subspaces are low probability regions that are densely scattered in the high dimensional representation space of DNNs. The properties of adversarial regions is considered as a key requirement for adversarial defense  \cite{ma2018characterizing}.

% \begin{figure}[ht]
% \vskip 0.15in
% \begin{center}
% \centerline{\includegraphics[width=\columnwidth]{Images/adv.png}}
% \caption{Examples of adversarial attacks crafted by the Projected Gradient Descent (PGD) to fool DNNs trained on medical image datasets: the fisrt row (Fundoscopy: (DR$=$diabetic retinopathy), the second row ( Chest X-Ray), and the third row (Dermoscopy). The left column indicates normal images, The middle column indicates adversarial perturbations, and The right column indicates adversarial images. The left bottom tag is the predicted class, and green/red indicates correct/wrong predictions \citep{ma2020understanding}}
% \label{fig:attack}
% \end{center}
% \vskip -0.15in
% \end{figure}
 
\section{Mahalanobis confidence score}

One of the limitations of both ODIN and LID is that they are designed for either OOD or adversarial corruption but not for both. More recently, \citet{lee2018simple} proposed a simple yet effective method for detecting both OOD samples and adversarial samples.

Mahalanobis distance-based confidence score is a class-conditional anomaly detection method, motivated by classification prediction confidence \citep{kamoi2020mahalanobis}. 

Let us consider a dataset $\mathcal{D}=\{ (\mathbf{x}_n,y_n )\}_{n=1}^N$ with input-label pairs. The labels belong to one of the classes $\{1,\ldots, C\}$. For malara parasite detection, labels are either parasitized or healthy. For a deep neural network, $f_\phi$, with parameters $\phi$, we consider a pre-trained softmax classifier,
\begin{equation}
p_{\theta}(y=c|\mathbf{x})=\frac{\exp(\mathbf{w}_c^\top f_{\phi}(\mathbf{x}))}{\sum_{c'}\exp(\mathbf{w}_{c'}^\top f_{\phi}(\mathbf{x}))}.
\end{equation}

For each class, we define a multivariate Gaussian  distribution,
$p (f_\phi(\mathbf{x})|y = c)=\mathcal{N} (f_\phi(\mathbf{x})| \bm{\mu}_c,\Sigma)$ with class mean $\bm{\mu}_c$ and pooled-covariance $\Sigma$. This way, we compute the empirical statistics $(\hat{\bm{\mu}}_1, \hat{\bm{\mu}}_2, \ldots, \hat{\bm{\mu}}_C, \hat{\Sigma})$ from the training dataset $\mathcal{D}$. With these statistics, the closest class $\tilde{c}$ to a query input $\mathbf{x}_*$ can be computed using the Mahalanobis distance,
\begin{equation}
\tilde{c} = \min_{c \in \{1,2,\ldots,C \}}  \sqrt{(f(\mathbf{x}_*) - \hat{\bm{\mu}}_c)^\top \hat{\Sigma}^{-1}(f(\mathbf{x}_*) - \hat{\bm{\mu}}_c) }.
\end{equation}

Following \citet{liang2017principled}, a controlled noise $\epsilon$ is added to the input,
\begin{equation}
{\tilde{\mathbf{x}}}_* = \mathbf{x}_* - \epsilon \cdot \mathrm{sign} \big(\nabla_\mathbf{x}(f(\mathbf{x}_*)- \bm{\mu}_{\tilde{c}})^\top \hat{\Sigma}^{-1} (f(\mathbf{x}_*)-\bm{\hat{\mu}}_{\tilde{c}} ) \big),
\end{equation}
for better calibration. Then, we can compute the confidence score,
\begin{equation}
M(\tilde{\mathbf{x}}_*) = \max_{c \in \{1,2,\ldots,C \}}  -(f(\tilde{\mathbf{x}}_*) - \hat{\bm{\mu}}_c)^\top \hat{\Sigma}^{-1}(f(\tilde{\mathbf{x}}_*) - \hat{\bm{\mu}}_c).
\end{equation}
By doing this for all $l=\{1,\ldots,L\}$ layers of the neural network with weights $\alpha_l$ of the logistic regression classifier (separately trained for each layer on a validation dataset \cite{ma2018characterizing}), we can compute the overall score $M^*(\mathbf{x_*}) = \sum_{l=1}^L \alpha_l M_l(\mathbf{x_*})$. For a given threshold $\rho$, the query samples, $\mathbf{x_*}$, is \emph{in-distribution}, if $M^*(\mathbf{x_*}) \geq \rho$.

\section{Experiments}

In  our  experiments,  we used  a publicly accessible and annotated malaria dataset of healthy and infected blood smear images \footnote{ \url{https://lhncbc.nlm.nih.gov/publication/pub9932}}. It contains $13,779$ parasitized and $13,779$  uninfected cell images. We split the dataset into $60:10:30$ for train, validation, and test datasets, respectively. We resized  the images to $125\times 125$ pixels and normalized them to assist in faster convergence. To prevent over-fitting and to account for possible variations in photomicroscopy, we have applied  data augmentation techniques such as rotation, shearing, translation, and zooming. For OOD, another malaria dataset consisting of $22,046$ was used\footnote{\url{https://github.com/shriyakabra97/malaria-parasite-detection}}.

\begin{figure}[ht]
\vskip 0.15in
\begin{center}
\centerline{\includegraphics[width=\columnwidth]{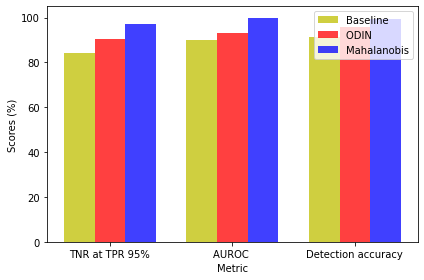}}
\caption{Robustness against \emph{out-of-distribution} samples: ResNet-18}
\label{fig:rp}
\end{center}
\vskip -0.15in
\end{figure}

\begin{figure}[ht]
\vskip 0.15in
\begin{center}
\centerline{\includegraphics[width=\columnwidth]{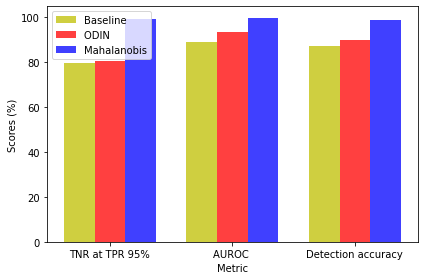}}
\caption{Robustness against \emph{out-of-distribution} samples: VGG-19}
\label{fig:vp}
\end{center}
\vskip -0.15in
\end{figure}

\begin{table*} []
		\caption{Robustness against \emph{adversarial samples}: a comparison of the performance of LID and Mahalanobis (proposed) towards detecting adversarial test samples generated from  malaria image datasets.}
			\centering
	\resizebox{\textwidth}{!}{\begin{tabular}{cccccc|cccccc}
			%\cline{1-10}
			\toprule
			{\bf Model} & {\bf Metric}& \multicolumn{4}{ c}{\bf LID} & \multicolumn{4}{ c}{\bf Mahalanobis} \\ \cmidrule{3-6} \cmidrule{7-10}
			\multicolumn{1}{ l  }{} &
			\multicolumn{1}{ l }{} & FGSM & BIM & DeepFool&CW&FGSM&BIM&DeepFool&CW    \\ \midrule %\cline{1-10} 
			\multicolumn{1}{ l  }{{\bf VGG-19}}                        &
			\multicolumn{1}{ l }{TNR at TPR 95\%} & 99.96& 96.87& 74.48&75.93&100.00&100.00 &75.96     &98.11          \\ %\cline{2-10}
			\multicolumn{1}{ l  }{}                        &
			\multicolumn{1}{ l }{AUROC} & 97.30 &  96.68
			& 78.01 &89.64&99.98&99.68&83.56& 99.35  \\ %\cline{2-10}		
			\multicolumn{1}{ l  }{}                        &                 
			\multicolumn{1}{ l }{Detection accuracy} &99.41&90.46&46.05&72.96&   99.98&99.99&61.95 &97.51  \\ \midrule%\cline{1-10}
			\multicolumn{1}{ l  }{{\bf ResNet-18} } &
			\multicolumn{1}{ l }{TNR at TPR 95\%} &96.84 &  95.61 & 63.59  &73.09 & 99.99 & 97.98&76.22 
			&98.90 
			\\ %\cline{2-10}
			\multicolumn{1}{ l  }{}                        &
			\multicolumn{1}{ l }{AUROC} & 97.30 &  96.68
			& 78.01 &89.64&99.98&99.68&83.56& 99.35  \\ %\cline{2-10}
			\multicolumn{1}{ l  }{}                        &
			\multicolumn{1}{ l }{Detection accuracy} &97.02 &97.65 & 49.56&84.66&99.75&99.95&64.95&98.05    \\ \bottomrule%\cline{1-10}
	\end{tabular}}
	\label{table2}
\end{table*}
\begin{figure}[ht]
\vskip 0.15in
\begin{center}
\centerline{\includegraphics[width=\columnwidth]{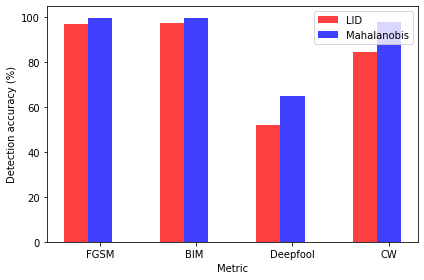}}
\caption{A comparison of performance between LID and Mahalanobis distance-based confidence score for ResNet-18 pre-trained model.}
\label{fig:rp2}
\end{center}
\vskip -0.15in
\end{figure}

\begin{figure}[ht]
\vskip 0.15in
\begin{center}
\centerline{\includegraphics[width=\columnwidth]{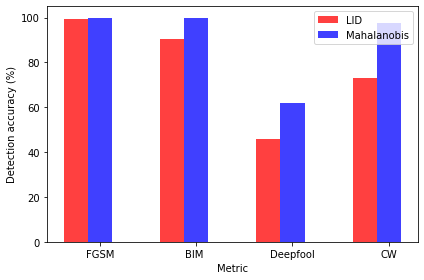}}
\caption{A comparison of detection performance between LID and Mahalanobis distance-based confidence score for VGG-19 pre-trained model}
\label{fig:vp2}
\end{center}
\vskip -0.15in
\end{figure}

For OOD and adversarial samples detection, the suggested method to improve the robustness of DL models was evaluated on both VGG-19 and  ResNet-18 using a threshold-based detector. We evaluate the models  with the following metrics: the true negative rate (TNR) at 95\% true positive rate (TPR), the area under the receiver operating characteristic (AUROC) curve, the area under the precision-recall (AUPR) curve, and the detection accuracy. The Mahalanobis confidence score was compared with the baseline method and state-of-the-art ODIN for OOD samples. It was also compared with the state-of-the-art LID toward adversarial samples detection. Comparing with the baseline method, ODIN and LID, as shown in Table~\ref{table2} and Figures~\ref{fig:rp}, \ref{fig:vp},\ref{fig:rp2},and \ref{fig:vp2}, the proposed approach outperforms on detecting abnormal samples.

As the Mahalanobis distance-based score  method outperforms for the tasks of detecting OOD samples and  adversarial samples, it can serve as a diagnostic framework for evaluating deep neural networks, as it is able to reveal  their potentially non-obvious vulnerabilities and reliability. Such frameworks help to ensure that deep neural networks are effective, secure, and easy to deploy in a  broad range of medical imaging applications beyond malaria detection. Our future work will extend this framework to test medical images under various lighting and other possible sources of corruption. We envision this, in the long-term, will enable low-cost, yet reliable, imaging procedures. 

\section*{Broader impact statement}
Broadly, our research is a step towards developing robust semi-automated medical image analysis techniques. Specifically, we focus on ensuring that malaria detection procedures are reliable enough before deployment. It, in the long-term, will help low-income countries to eradicate malaria. These systems, however, need to be rigorously validated before deploying in medical facilities. 

\nocite{langley00}

\bibliography{main}

\begin{thebibliography}{31}
\providecommand{\natexlab}[1]{#1}
\providecommand{\url}[1]{\texttt{#1}}
\expandafter\ifx\csname urlstyle\endcsname\relax
  \providecommand{\doi}[1]{doi: #1}\else
  \providecommand{\doi}{doi: \begingroup \urlstyle{rm}\Url}\fi

\bibitem[Ardila et~al.(2019)Ardila, Kiraly, Bharadwaj, Choi, Reicher, Peng,
  Tse, Etemadi, Ye, Corrado, et~al.]{ardila2019end}
Ardila, D., Kiraly, A.~P., Bharadwaj, S., Choi, B., Reicher, J.~J., Peng, L.,
  Tse, D., Etemadi, M., Ye, W., Corrado, G., et~al.
\newblock End-to-end lung cancer screening with three-dimensional deep learning
  on low-dose chest computed tomography.
\newblock \emph{Nature medicine}, 25\penalty0 (6):\penalty0 954--961, 2019.

\bibitem[Asgharnezhad et~al.(2020)Asgharnezhad, Shamsi, Alizadehsani, Khosravi,
  Nahavandi, Sani, and Srinivasan]{asgharnezhad2020objective}
Asgharnezhad, H., Shamsi, A., Alizadehsani, R., Khosravi, A., Nahavandi, S.,
  Sani, Z.~A., and Srinivasan, D.
\newblock Objective evaluation of deep uncertainty predictions for covid-19
  detection.
\newblock \emph{arXiv preprint arXiv:2012.11840}, 2020.

\bibitem[Carlini \& Wagner(2017)Carlini and Wagner]{carlini2017adversarial}
Carlini, N. and Wagner, D.
\newblock Adversarial examples are not easily detected: Bypassing ten detection
  methods.
\newblock In \emph{Proceedings of the 10th ACM Workshop on Artificial
  Intelligence and Security}, pp.\  3--14, 2017.

\bibitem[Chen et~al.(2020)Chen, Wu, Liang, Jha, et~al.]{chen2020robust}
Chen, J., Wu, X., Liang, Y., Jha, S., et~al.
\newblock Robust out-of-distribution detection in neural networks.
\newblock \emph{arXiv preprint arXiv:2003.09711}, 2020.

\bibitem[Cho et~al.(2020)Cho, Sun, Mun, Kim, Kim, Cho, Youn, Kim, and
  Chung]{cho2020dermatologist}
Cho, S.~I., Sun, S., Mun, J.-H., Kim, C., Kim, S., Cho, S., Youn, S., Kim,
  H.~C., and Chung, J.
\newblock Dermatologist-level classification of malignant lip diseases using a
  deep convolutional neural network.
\newblock \emph{British Journal of Dermatology}, 182\penalty0 (6):\penalty0
  1388--1394, 2020.

\bibitem[{Dong} et~al.(2017){Dong}, {Jiang}, {Shen}, {David Pan}, {Williams},
  {Reddy}, {Benjamin}, and {Bryan}]{7897215}
{Dong}, Y., {Jiang}, Z., {Shen}, H., {David Pan}, W., {Williams}, L.~A.,
  {Reddy}, V. V.~B., {Benjamin}, W.~H., and {Bryan}, A.~W.
\newblock Evaluations of deep convolutional neural networks for automatic
  identification of malaria infected cells.
\newblock In \emph{2017 IEEE EMBS International Conference on Biomedical Health
  Informatics (BHI)}, pp.\  101--104, 2017.

\bibitem[Emmott et~al.(2016)Emmott, Das, Dietterich, Fern, and
  Wong]{emmott2016metaanalysis}
Emmott, A., Das, S., Dietterich, T., Fern, A., and Wong, W.-K.
\newblock A meta-analysis of the anomaly detection problem, 2016.

\bibitem[Esteva et~al.(2017)Esteva, Kuprel, Novoa, Ko, Swetter, Blau, and
  Thrun]{esteva2017dermatologist}
Esteva, A., Kuprel, B., Novoa, R.~A., Ko, J., Swetter, S.~M., Blau, H.~M., and
  Thrun, S.
\newblock Dermatologist-level classification of skin cancer with deep neural
  networks.
\newblock \emph{nature}, 542\penalty0 (7639):\penalty0 115--118, 2017.

\bibitem[Fawzi et~al.(2016)Fawzi, Moosavi-Dezfooli, and
  Frossard]{NIPS2016_6331}
Fawzi, A., Moosavi-Dezfooli, S.-M., and Frossard, P.
\newblock Robustness of classifiers: from adversarial to random noise.
\newblock In Lee, D.~D., Sugiyama, M., Luxburg, U.~V., Guyon, I., and Garnett,
  R. (eds.), \emph{Advances in Neural Information Processing Systems 29}, pp.\
  1632--1640. Curran Associates, Inc., 2016.

\bibitem[Goodfellow et~al.(2014)Goodfellow, Pouget-Abadie, Mirza, Xu,
  Warde-Farley, Ozair, Courville, and Bengio]{goodfellow2014generative}
Goodfellow, I., Pouget-Abadie, J., Mirza, M., Xu, B., Warde-Farley, D., Ozair,
  S., Courville, A., and Bengio, Y.
\newblock Generative adversarial nets.
\newblock In \emph{Advances in neural information processing systems}, pp.\
  2672--2680, 2014.

\bibitem[Goodfellow et~al.(2015)Goodfellow, Shlens, and
  Szegedy]{goodfellow2015explaining}
Goodfellow, I.~J., Shlens, J., and Szegedy, C.
\newblock Explaining and harnessing adversarial examples.
\newblock In \emph{The International Conference on Learning Representations},
  2015.

\bibitem[Haenssle et~al.(2018)Haenssle, Fink, Schneiderbauer, Toberer, Buhl,
  Blum, Kalloo, Hassen, Thomas, Enk, et~al.]{haenssle2018man}
Haenssle, H.~A., Fink, C., Schneiderbauer, R., Toberer, F., Buhl, T., Blum, A.,
  Kalloo, A., Hassen, A. B.~H., Thomas, L., Enk, A., et~al.
\newblock Man against machine: diagnostic performance of a deep learning
  convolutional neural network for dermoscopic melanoma recognition in
  comparison to 58 dermatologists.
\newblock \emph{Annals of Oncology}, 29\penalty0 (8):\penalty0 1836--1842,
  2018.

\bibitem[Hendrycks \& Gimpel(2016)Hendrycks and Gimpel]{hendrycks2016baseline}
Hendrycks, D. and Gimpel, K.
\newblock A baseline for detecting misclassified and out-of-distribution
  examples in neural networks.
\newblock \emph{arXiv preprint arXiv:1610.02136}, 2016.

\bibitem[Kamoi \& Kobayashi(2020)Kamoi and Kobayashi]{kamoi2020mahalanobis}
Kamoi, R. and Kobayashi, K.
\newblock Why is the mahalanobis distance effective for anomaly detection?
\newblock \emph{arXiv preprint arXiv:2003.00402}, 2020.

\bibitem[Kurakin et~al.(2016)Kurakin, Goodfellow, and
  Bengio]{kurakin2016adversarial}
Kurakin, A., Goodfellow, I., and Bengio, S.
\newblock Adversarial examples in the physical world.
\newblock \emph{arXiv preprint arXiv:1607.02533}, 2016.

\bibitem[Laves et~al.(2020)Laves, Ihler, Fast, Kahrs, and
  Ortmaier]{laves2020well}
Laves, M.-H., Ihler, S., Fast, J.~F., Kahrs, L.~A., and Ortmaier, T.
\newblock Well-calibrated regression uncertainty in medical imaging with deep
  learning.
\newblock In \emph{Medical Imaging with Deep Learning}, pp.\  393--412. PMLR,
  2020.

\bibitem[Lee et~al.(2018)Lee, Lee, Lee, and Shin]{lee2018simple}
Lee, K., Lee, K., Lee, H., and Shin, J.
\newblock A simple unified framework for detecting out-of-distribution samples
  and adversarial attacks.
\newblock In \emph{Advances in Neural Information Processing Systems}, pp.\
  7167--7177, 2018.

\bibitem[Liang et~al.(2017)Liang, Li, and Srikant]{liang2017principled}
Liang, S., Li, Y., and Srikant, R.
\newblock Principled detection of out-of-distribution examples in neural
  networks.
\newblock \emph{arXiv preprint arXiv:1706.02690}, 2017.

\bibitem[Liang et~al.(2018)Liang, Li, and Srikant]{liang2017enhancing}
Liang, S., Li, Y., and Srikant, R.
\newblock Enhancing the reliability of out-of-distribution image detection in
  neural networks.
\newblock In \emph{The International Conference on Learning Representations},
  2018.

\bibitem[Liang et~al.(2016)Liang, Powell, Ersoy, Poostchi, Silamut,
  Palaniappan, Guo, Hossain, Antani, Maude, Huang, Jaeger, and
  Thoma]{Liang2016CNNbasedIA}
Liang, Z., Powell, A., Ersoy, I., Poostchi, M., Silamut, K., Palaniappan, K.,
  Guo, P., Hossain, M.~A., Antani, S.~K., Maude, R.~J., Huang, X., Jaeger, S.,
  and Thoma, G.~R.
\newblock Cnn-based image analysis for malaria diagnosis.
\newblock \emph{2016 IEEE International Conference on Bioinformatics and
  Biomedicine (BIBM)}, pp.\  493--496, 2016.

\bibitem[Ma et~al.(2018)Ma, Li, Wang, Erfani, Wijewickrema, Schoenebeck, Song,
  Houle, and Bailey]{ma2018characterizing}
Ma, X., Li, B., Wang, Y., Erfani, S.~M., Wijewickrema, S., Schoenebeck, G.,
  Song, D., Houle, M.~E., and Bailey, J.
\newblock Characterizing adversarial subspaces using local intrinsic
  dimensionality.
\newblock In \emph{The International Conference on Learning Representations},
  2018.

\bibitem[Moosavi-Dezfooli et~al.(2016)Moosavi-Dezfooli, Fawzi, and
  Frossard]{moosavi2016deepfool}
Moosavi-Dezfooli, S.-M., Fawzi, A., and Frossard, P.
\newblock Deepfool: a simple and accurate method to fool deep neural networks.
\newblock In \emph{Proceedings of the IEEE conference on computer vision and
  pattern recognition}, pp.\  2574--2582, 2016.

\bibitem[Nitsch et~al.(2021)Nitsch, Itkina, Senanayake, Nieto, Schmidt,
  Siegwart, Kochenderfer, and Cadena]{nitsch2020out}
Nitsch, J., Itkina, M., Senanayake, R., Nieto, J., Schmidt, M., Siegwart, R.,
  Kochenderfer, M.~J., and Cadena, C.
\newblock Out-of-distribution detection for automotive perception.
\newblock In \emph{24th IEEE International Conference on Intelligent
  Transportation Systems (ITSC)}, 2021.
\newblock \doi{arXiv:2011.01413}.

\bibitem[Rajaraman et~al.(2018{\natexlab{a}})Rajaraman, Antani, Poostchi,
  Silamut, Hossain, Maude, Jaeger, and Thoma]{Dipanjan}
Rajaraman, S., Antani, S.~K., Poostchi, M., Silamut, K., Hossain, M.~A., Maude,
  R.~J., Jaeger, S., and Thoma, G.~R.
\newblock Pre-trained convolutional neural networks as feature extractors
  toward improved malaria parasite detection in thin blood smear images.
\newblock \emph{PeerJ}, 6:\penalty0 e4568, 2018{\natexlab{a}}.

\bibitem[Rajaraman et~al.(2018{\natexlab{b}})Rajaraman, Antani, Poostchi,
  Silamut, Hossain, Maude, Jaeger, and Thoma]{rajaraman2018pre}
Rajaraman, S., Antani, S.~K., Poostchi, M., Silamut, K., Hossain, M.~A., Maude,
  R.~J., Jaeger, S., and Thoma, G.~R.
\newblock Pre-trained convolutional neural networks as feature extractors
  toward improved malaria parasite detection in thin blood smear images.
\newblock \emph{PeerJ}, 6:\penalty0 e4568, 2018{\natexlab{b}}.

\bibitem[Ravendran et~al.(2015)Ravendran, de~Silva, and
  Senanayake]{ravendran2015moment}
Ravendran, A., de~Silva, K. R.~T., and Senanayake, R.
\newblock Moment invariant features for automatic identification of critical
  malaria parasites.
\newblock In \emph{2015 IEEE 10th International Conference on Industrial and
  Information Systems (ICIIS)}, pp.\  474--479. IEEE, 2015.

\bibitem[Senanayake et~al.(2016)Senanayake, O'Callaghan, and
  Ramos]{senanayake2016predicting}
Senanayake, R., O'Callaghan, S., and Ramos, F.
\newblock Predicting spatio-temporal propagation of seasonal influenza using
  variational gaussian process regression.
\newblock In \emph{Proceedings of the AAAI Conference on Artificial
  Intelligence}, volume~30, 2016.

\bibitem[Silva et~al.(2013)Silva, Wijesundara, and
  Senanayake]{silva2013computer}
Silva, A. D.~N., Wijesundara, M.~N., and Senanayake, R.
\newblock Computer controlled digital microscope with photomicrograph
  enhancement.
\newblock In \emph{2013 International Conference of Information and
  Communication Technology (ICoICT)}, pp.\  44--47. IEEE, 2013.

\bibitem[Wongsrichanalai et~al.(2007)Wongsrichanalai, Barcus, Muth,
  Sutamihardja, and Wernsdorfer]{wongsrichanalai2007review}
Wongsrichanalai, C., Barcus, M.~J., Muth, S., Sutamihardja, A., and
  Wernsdorfer, W.~H.
\newblock A review of malaria diagnostic tools: microscopy and rapid diagnostic
  test (rdt).
\newblock \emph{The American journal of tropical medicine and hygiene},
  77\penalty0 (6\_Suppl):\penalty0 119--127, 2007.

\bibitem[{\relax World Health Organization}()]{WHO}
{\relax World Health Organization}.
\newblock World malaria report 2019.
\newblock https://www.who.int/news-room/fact-sheets/detail/malaria.
\newblock Accessed: 2020-05-25.

\bibitem[Yang et~al.(2019)Yang, Poostchi, Yu, Zhou, Silamut, Yu, Maude, Jaeger,
  and Antani]{yang2019deep}
Yang, F., Poostchi, M., Yu, H., Zhou, Z., Silamut, K., Yu, J., Maude, R.~J.,
  Jaeger, S., and Antani, S.
\newblock Deep learning for smartphone-based malaria parasite detection in
  thick blood smears.
\newblock \emph{IEEE journal of biomedical and health informatics}, 24\penalty0
  (5):\penalty0 1427--1438, 2019.

\end{thebibliography}
\bibliographystyle{icml2021}


%%%%%%%%%%%%%%%%%%%%%%%%%%%%%%%%%%%%%%%%%%%%%%%%%%%%%%%%%%%%%%%%%%%%%%%%%%%%%%%
%%%%%%%%%%%%%%%%%%%%%%%%%%%%%%%%%%%%%%%%%%%%%%%%%%%%%%%%%%%%%%%%%%%%%%%%%%%%%%%

\end{document}